\documentclass{vldb}
\usepackage{graphicx}
\usepackage{balance}  
\usepackage[dvipsnames]{xcolor}
\usepackage{comment}
\usepackage{xspace}
\usepackage{cite}
\usepackage{enumitem}
\usepackage{algorithm}
\usepackage[noend]{algpseudocode}

\newcommand{\revision}[1]{{\color{black}{#1}}}

\newcommand{\sugato}[1]{{\color{blue}sugato: {#1}}}
\newcommand{\new}[1]{{{#1}}}
\newcommand{\ch}[1]{{{#1}}}

\newcommand{\orgres}{organizational resource\xspace}

\newcommand{\orgress}{organizational resources\xspace}
\newcommand{\OrgRess}{Organizational Resources\xspace}
\newcommand{\Orgress}{Organizational resources\xspace}

\newcommand{\splitarch}{split architecture\xspace}

\newcommand{\hardlinks}{hard links\xspace}
\newcommand{\softlinks}{soft links\xspace}

\newcommand{\Softlinks}{Soft links\xspace}

\newcommand{\minihead}[1]{{\vspace{.45em}\noindent\textbf{#1.} }}
\newcommand{\miniheadit}[1]{{\vspace{.45em}\noindent\textit{#1.} }}

\newcommand{\eat}[1]{}

\vldbTitle{Leveraging \OrgRess to Adapt Models to New Data Modalities}
\vldbAuthors{Sahaana Suri,  Raghuveer Chanda, Neslihan Bulut, Pradyumna Narayana, Yemao Zeng, Peter Bailis, Sugato Basu, Girija Narlikar, Christopher R\'e, Abishek Sethi}
\vldbDOI{https://doi.org/10.14778/3415478.3415559}
\vldbVolume{13}
\vldbNumber{12}
\vldbYear{2020}

\begin{document}


\title{Leveraging \OrgRess to Adapt Models to New Data Modalities}


\author{
\alignauthor
Sahaana Suri$^\dag$\thanks{This work was done while the author was at Google},  Raghuveer Chanda, Neslihan Bulut, Pradyumna Narayana, Yemao Zeng\\
Peter Bailis$^\dag$, Sugato Basu, Girija Narlikar, Christopher R\'e$^\dag$, Abishek Sethi \\
\vspace{0.25em}
\affaddr{Google, Stanford$^\dag$}
\email{\{sahaana, pbailis, chrismre\}@stanford.edu,\\
\{rachanda, neslihanbulut, pradyn, yemao, sugato, girijan, abishek\}@google.com}
}

\maketitle

\begin{abstract}
As applications in large organizations evolve, the machine learning (ML) models that power them must adapt the same predictive tasks to newly arising data modalities (e.g., a new video content launch in a social media application requires existing text or image models to extend to video). 
To solve this problem, organizations typically create ML pipelines from scratch. 
However, this fails to utilize the domain expertise and data they have cultivated from developing tasks for existing modalities.
We demonstrate how \emph{\orgress}, in the form of aggregate statistics, knowledge bases, and existing services that operate over related tasks, enable teams to construct a common feature space that connects new and existing data modalities.
This allows teams to apply methods for data curation (e.g., weak supervision and label propagation) and model training (e.g., forms of multi-modal learning) across these different data modalities.
We study how this use of \orgress composes at production scale in over \ch{5} classification tasks at Google, and demonstrate how it reduces the time needed to develop models for new modalities from months \revision{to weeks or days.} 

\eat{
As applications in large organizations evolve, the machine learning (ML) models that power them must adapt the same predictive tasks to newly arising data modalities (e.g., a new video content launch in a social media application requires existing text or image models to extend to video). To solve this problem, organizations typically create ML pipelines from scratch. However, this fails to utilize the domain expertise and data they have cultivated from developing tasks for existing modalities. We demonstrate how organizational resources, in the form of aggregate statistics, knowledge bases, and existing services that operate over related tasks, enable teams to construct a common feature space that connects new and existing data modalities. This allows teams to apply methods for data curation (e.g., weak supervision and label propagation) and model training (e.g., forms of multi-modal learning) across these different data modalities. We study how this use of organizational resources composes at production scale in over five classification tasks at Google, and demonstrate how it reduces the time needed to develop models for new modalities from months to weeks or days.
}
\end{abstract}

\section{Introduction}

Large organizations that rely on machine learning (ML) models for tasks such as content and event classification often adapt existing models to new data modalities to perform the \emph{same} predictive tasks over these new modalities.
Consider the following example based on a Google team.
\\

\emph{A content moderation team uses an ML pipeline to flag policy-violating user posts on a social media application (e.g., \new{harmful speech, spam, or sensitive content}).
While the application initially supported text- and image-based posts, the application will soon support video-based posts.
The moderators must thus classify new video posts for the same violations as the text and image posts. 
}
\\

\begin{figure}
  \includegraphics[width=\linewidth]{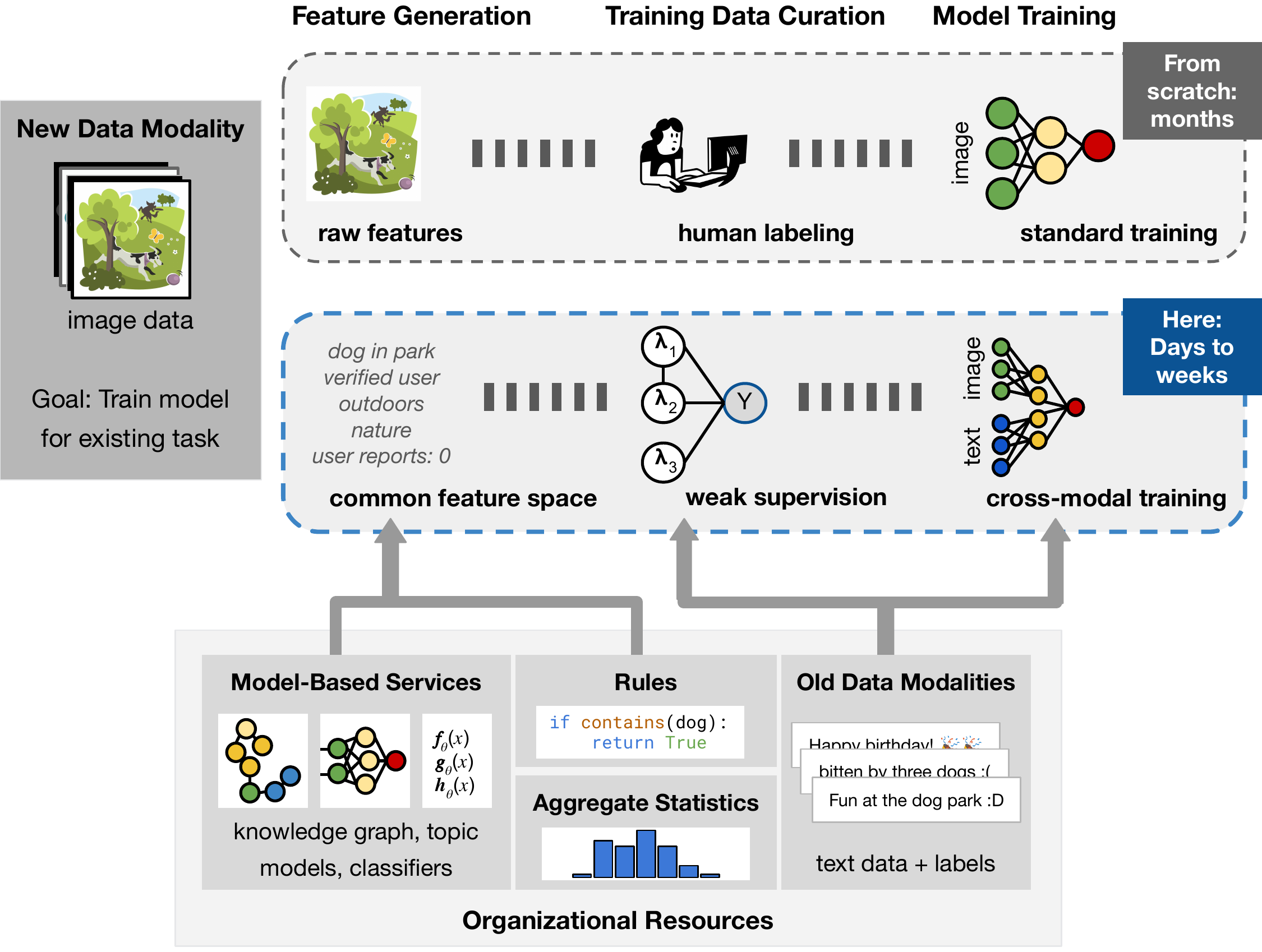}
  \caption{Traditional means of cross-modal adaptation take months to deploy a new model (top), whereas leveraging organizational resources at each step (bottom) shrinks that time to days \revision{or weeks}.}
  \label{fig:sys_overview}
\end{figure}

We refer to this process of adapting models to new data modalities as \emph{cross-modal adaptation} (a form of transductive transfer learning~\cite{transferlearning1}). 
Existing work in cross-modal adaptation assumes points across data modalities are easily or directly linked (e.g., captions directly linked to images, or clinical notes to lab results) to leverage zero-shot learning~\cite{zscm2,zscm3} or weak supervision~\cite{crossmodalws}.
However, in our environment, such direct connections do not often exist, resulting in a \emph{modality gap}. For example, new video posts may not contain any descriptive content (e.g., text summaries), and may bear no relation to a users' previous posts. 

In our experience, production cross-modal deployments typically fail to address the modality gap and instead build a standard ML pipeline from scratch, following a three-step \emph{\splitarch}~\cite{ML} (Figure~\ref{fig:sys_overview}, top): 
\begin{enumerate}
    \itemsep-.2em
    \item Feature Generation (optional): featurize data of the new modality.
    \item Training Data Curation: label data of the new modality (often via sampling and human labeling).
    \item Model Training: train a model on the labeled training data of the new modality.
\end{enumerate}

This classic ML approach can take months to complete and leads to a set of disparate models with both their own feature spaces and training data dependent on data modality. 
In our moderation example, as direct translations of policy violations are unclear when moving from a static to sequential modality, the team develops models from scratch.
They rely on neural network models that elide explicit feature generation, and spend weeks to months labeling video data to train models to identify policy violations.
While this procedure fails to leverage previously developed expertise, the general and modular \splitarch allows small, yet diverse engineering teams to easily deploy and monitor models for new tasks~\cite{techdebt,mlinfra}.
Thus, a natural question arises: {\em is it possible to bootstrap data-limited cross-modal adaptation pipelines by augmenting the existing \splitarch?}

\begin{figure}
  \includegraphics[width=\linewidth]{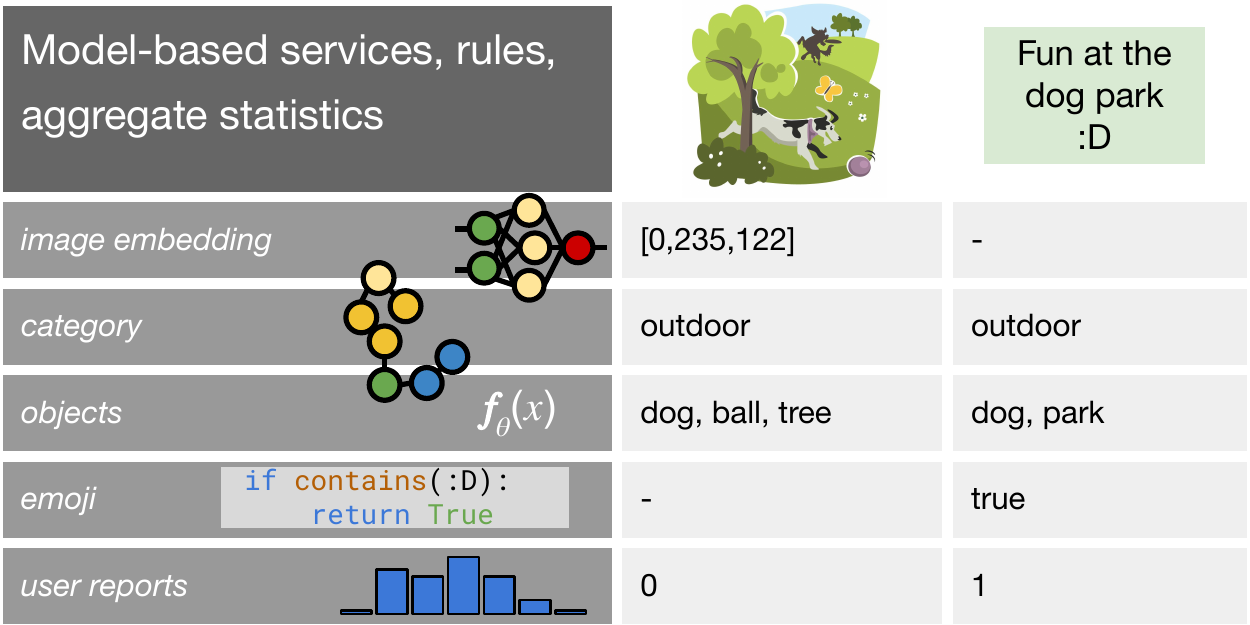}
  \caption{Using \orgress to create a common, structured feature space across data modalities for a post moderation task. Image and text data share three features: category (topic model), objects present (classifier for related task), and the number of times the user posting the content has been reported (aggregate statistic).}
  \label{fig:feature_gen}
\end{figure}

We find that despite the modality gap, we can leverage auxiliary data sources to connect points across data modalities. We refer to these auxiliary links as \emph{organizational resources}, which we exploit to boost the effectiveness of each pipeline step for cross-modal adaptation. 
Organizations cultivate public and proprietary resources in the form of tools or services that take existing data points as input, and return features, metadata or statistics that describe them (Figure~\ref{fig:feature_gen}).
Even in the absence of custom classifiers for a new modality (e.g., sensitive content detection over video posts), it is still possible to apply existing resources from across an organization (e.g., knowledge-graph querying tools for videos).
This holds across large enterprises~\cite{OR_aws,OR_googlecloud}, academic labs~\cite{OR_stanfordnlp} and medical institutions~\cite{crossmodalws}.
As a result, the data management research community has developed systems and algorithms for systematically leveraging these resources for ML tasks~\cite{modeldb,snorkel,drybell,columnml,mmdbms1}.
In this work, we demonstrate how to leverage \orgress to bridge the modality gap and improve each of the three steps of the \splitarch as follows (Figure~\ref{fig:sys_overview}, bottom):


\minihead{Feature Generation: use \orgress to create common features} The first step to overcome the modality gap is to construct features that are common between modalities.
We find that a straightforward, yet effective means of achieving this is by identifying \orgress that can transform data points to representations common across modalities (e.g., in a topic modeling system that applies to text and image, the common representation is the topic of the content). 
In this way, we view \orgress as a library of feature transformations: we pass data points of different modalities into these services and compose their outputs to form rich shared feature spaces (Figure~\ref{fig:feature_gen}).
These common features lay the foundation for improvements in the remaining two \splitarch steps.

\minihead{Training Data Curation: use weak supervision with label propagation} 
Given the common feature space, one approach to cross-modal adaptation is to train a model with labeled data from existing modalities using just the features shared between modalities. 
We can then perform inference over the new data modality using the shared features.
We find that this baseline performs worse than training a model on the target modality with respect to AUPRC, likely due to distribution differences in the feature spaces.
Thus, we still require labeled training data in the new modality.

To mitigate the cost of obtaining hand-labeling training data, organizations have leveraged weak supervision systems, such as Snorkel~\cite{drybell}, that use labeling functions (LFs) to programmatically label groups of data points.
However, we encounter three challenges in using weak supervision:

\miniheadit{Challenge 1: Curating Structured Data and a Development Set} LFs require users to specify predicates over their data that have both high precision and recall (e.g., \emph{if a text span contains the word SPAM, mark it as spam}). 
This is difficult in our setting for two reasons. 
First, specifying predicates over unstructured data like video and images is an active area of research~\cite{coral,dugong,crossmodalws} (e.g., it is difficult to easily specify what a policy-violating image looks like).
Second, to evaluate the performance of candidate LFs, we need a labeled development set.
To address these dual concerns, we leverage the common feature space we create in the first step.
As model outputs are frequently categorical and quantitative (e.g., output of an object detection routine), it is easy to define predicates over these features.
In addition, we can use labeled data of existing modalities as a development set.

\miniheadit{Challenge 2: Creating LFs Without Domain Expertise}
Domain experts typically construct LFs using task expertise.
However, in a production setting, experts may not be immediately available to develop LFs, and engineers often do not possess this expertise.
Existing methods to automatically generate LFs~\cite{snuba} are too costly to immediately integrate with existing workflows. 
As a result, we develop new techniques to automatically create LFs for our tasks. 
We leverage frequent itemset mining~\cite{fim} to more quickly and easily develop LFs.
Itemset mining automatically identifies feature values that occur more frequently in positive examples, which we can treat as LFs.
This method of LF generation enables us to mine our entire labeled data corpus for existing data modalities (tens of millions) in minutes.
In contrast, even domain experts are limited to manually examining much smaller data volumes. 
Thus, we find our method can be faster, cheaper, and perform better than domain-expert-curated LFs with respect to F1 score and coverage.

\miniheadit{Challenge 3: Finding Borderline Examples}
Weak supervision requires high precision and recall LFs that cover a majority of data points.
In our heavily class-imbalanced settings, we find that developing high-precision LFs to identify positive examples can be straightforward, but constructing rules to identify borderline positive and negative examples, which are crucial for recall and coverage, is difficult. 
In response, we use label propagation~\cite{labelprop} to augment our automatically mined LFs.
Label propagation detects data points in the new modality that are similar to labeled examples in the old modalities, with similarity defined using features in the common feature space.
Thus, we identify large volumes of negative examples and more candidate positives than with itemset mining, improving performance by up to 129$\times$ and \ch{1.25$\times$} with respect to recall and AUPRC, respectively.

\minihead{Model Training: combine data and label sources}  
Given the common features, we can leverage multi-modal techniques for model training that combine inputs from multiple data and label sources (e.g., data from new and existing modalities, human-generated labels, and labels from weak supervision).
We evaluate three techniques for combining the features for model training: by concatenating the features directly, concatenating embeddings independently learned for each data modality, and by projecting the new modality to an embedding learned using existing modalities.
We demonstrate that combining label and data modalities improves end-modal performance by up to \ch{1.63$\times$} in comparison to using any modality in isolation, and that simple feature concatenation outperforms the alternatives. 
\\

Overall, we make the following contributions in this paper:
 
\begin{itemize}
\item We outline how our augmented three-step split architecture addresses the production challenges in deploying, maintaining and evaluating pipelines for cross-modal adaptation with access to the limited but rich ecosystem of human resources present in a typical industrial team. We validate our design decisions using \ch{5} cross-modal classification workloads at Google.  
\item We demonstrate how using \orgress to augment the \splitarch enables us to develop cross-modal pipelines that obtain the same classification performance as using up to 750k fully supervised image data points by instead using 7.4M unlabeled image data points, and 25M previously hand-labeled text data points---decreasing the time to develop models for cross-modal adaptation from months to days \revision{or weeks}.
\item We develop a pipeline that overcomes the challenges of using weak supervision for cross-modal adaptation by automatically generating labeling functions up to \ch{1.87$\times$} faster than a domain expert, who must divide the task into days or weeks.
We also obtain increased performance with respect to coverage and F1 score. 
\end{itemize}

\section{Cross-Modal Adaptation}
\label{sec:background}

In this section, we define cross-modal adaptation and describe key challenges we overcome to develop a cross-modal pipeline in production.
\new{We divide the described challenges into three categories: contending with limited human resources, handling new data modalities, or uniting available \orgress.}
We then provide an overview of our system design.

 \subsection{Problem Statement}
\label{subsec:cross_modal}

Growing organizations increasingly support applications over multiple data modalities---products that may initially only support text must evolve to support richer modalities including image, videos, or animations (e.g., gifs). 
As organizations increasingly rely on machine learning (ML) models for content and event classification, they must therefore develop models to perform existing classification tasks over these new modalities as they arise.
We refer to this problem as \emph{cross-modal adaptation}:
our goal is to train a model for existing classification tasks over the new data modality as quickly as possible, when labeled data of the new modality is limited or nonexistent at the desired time of deployment. 

\new{
We assume a user can access \orgress to process given modalities and return structured (i.e., categorical or quantitative) outputs, and that new modalities provide additional means of conveying of information (i.e., are as rich or richer than existing modalities). 
While we construct examples based on adapting text and image tasks for video, our techniques apply to other commonly-processed modalities including audio signals, time series, point clouds, or network behavior in graphs.

}

\subsection{Cross-Modal Challenges}
\label{subsec:cm_challenges}
We highlight three challenges in cross-modal adaptation:

\minihead{[Human Resources] Labeling Rich Modalities}
\revision{For several learning tasks in production, labeling training data is a labor-intensive and time-consuming procedure, especially when facing large class imbalances~\cite{snorkel}.}
Referring to our example from the introduction, the team must sample hundreds to thousands of data points to find a few examples of sensitive content.
\revision{In our workloads, the cost of labeling richer data modalities is greater than that of existing modalities (e.g., manually classifying text is faster than viewing and classifying video). 
We only focus on this subset of cross-modal tasks, where reviewing new modalities is increasingly costly, and thus requires alternative labeling methodology.
}

\minihead{[Data Modalities] Bridging the Modality Gap}
Solutions for similar cross-modal problems (see Section~\ref{sec:relwork}) assume that other tasks have already been trained for the target modality~\cite{zscm1,zscm2,zscm3}, or data of different modalities are directly connected~\cite{crossmodalws}.
Examples of direct connections are images paired with captions, 2D projections of 3D point clouds, or clinical notes and lab results. 
The setting we consider often lacks these connections, resulting in a \emph{modality gap} between data points that we must bridge to leverage information and resources across modalities.

\minihead{[Organizational Resources] ~ Leveraging Resources Across Task, Data, and Label Source}
Organizations possess large amounts of data and expertise generated across existing tasks and data modalities (see Section~\ref{subsec:feat_gen}). 
In cross-modal adaptation, we must identify how to combine these information sources to train a high-performing end-model.

\subsection{Design Considerations in Production}
\label{subsec:design_considerations}

We outline four obstacles to the deployment and maintenance of cross-modal pipelines at scale.
\eat{While our focus is on cross-modal adaptation, these challenges apply to any large-scale, industrial ML deployment.}

\minihead{[Human Resources] Diverse Teams} A typical ML pipeline is developed by a rich ecosystem of humans across each pipeline step. 
Domain experts and human reviewers curate training data and validate the performance of ML models; systems engineers develop the architecture and infrastructure to easily deploy models; ML engineers train and tune models for deployment; and quality assurance engineers ensure that criteria for successful deployment are being met. 
A system for cross-modal adaptation must enable each individual to enter and exit at different steps of the pipeline to contribute their expertise, implying well-defined inputs and outputs for each step to decrease interference across roles.

\minihead{[Human Resources] Low Engineer-to-Pipeline Ratio} Machine learning teams, though rich in roles, are small in contrast to the hundreds---or thousands---of ML pipelines they must deploy and maintain. 
Each individual must perform their role as efficiently as possible in a manner that can scale to these large pipeline volumes.
A system for cross-modal adaptation must allow each individual to develop, deploy and manage their roles as easily and quickly as possible, implying a repeatable procedure that can be applied across as many of their tasks as possible.

\minihead{[Data Modalities] Rich Data Modalities} 
Richer data modalities (e.g., image or video as opposed to text) pose a challenge at each of the three \splitarch steps.
For feature generation, teams are allocated storage budgets that are sometimes insufficient to capture all incoming raw data.
\new{For curating training data, new modalities require integrated interfaces to enable human reviewers to select and evaluate data points.}
Finally, ML models must have low inference latency to be deployed in real-time, especially if user-facing; rich modalities are more expensive to process in raw feature space. 
A system for cross-modal adaptation must rely on efficient storage, classification, and inference over these rich modalities, to incur minimal overhead when these modalities are released in the application.

\minihead{[Organizational Resources] Diverse Info Sources}
Input features for various steps of the ML pipeline are drawn from services common across the organization (e.g., a common service to return the content of a post). 
Coordinating the input features of different models may require changes to system architectures if drawing from a new feature source, and necessitate population in advance.
As a result, adding new features to a deployed model is often a time-consuming process.
In addition, not all accessible features can be served at inference time due to the cost of extracting them from their respective data sources and systems. 
A system for cross-modal adaptation must be able to leverage features even if they cannot be deployed and served in production.

\begin{figure*}
  \includegraphics[width=\linewidth]{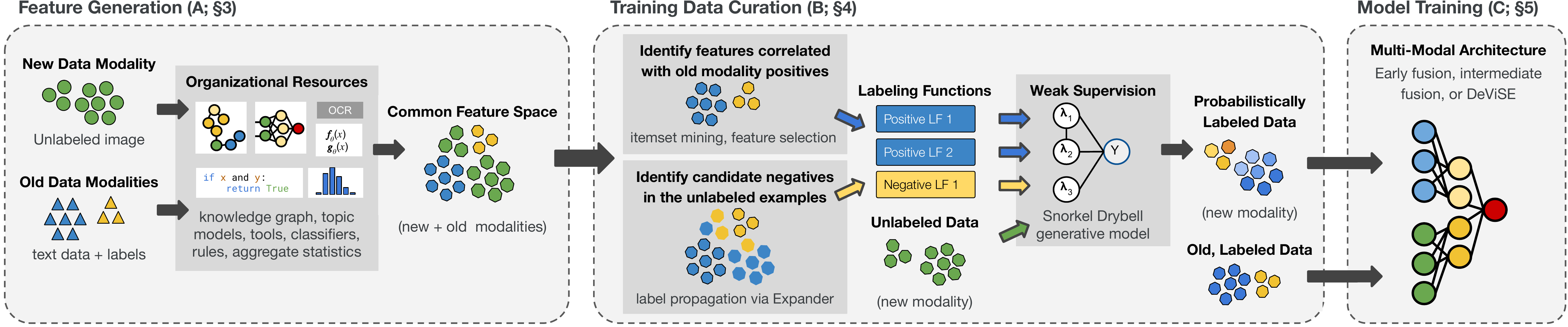}
  \caption{The steps of our system for cross-modal adaptation. Data of both new (unlabeled) and old (labeled) modalities are first transformed to a common feature space (A). This transformation enables us to use weak supervision to curate labeled training data in the new modality (B). Finally, labeled data of all modalities are jointly trained using a multi-modal architecture (C).}
  \label{fig:system}
\end{figure*}

\subsection{Solution Overview}

We develop a system for cross-modal adaptation that addresses both sets of challenges (Figure~\ref{fig:system}). 
In this section, we describe our system architecture and pipeline steps.

\subsubsection{Architecture}
We preserve the canonical \splitarch to ensure ease of management and deployment in light of the first two production challenges (\S~\ref{subsec:design_considerations}): diverse teams and low engineer-to-ML-pipeline ratios (Figure~\ref{fig:system}).
Having different architectures for each task, data modality, and label source is challenging to manage. 
A modular and general architecture that is reused across task-data-label configuration makes it easy for even a single engineer to maintain several ML pipelines. 

In addition, this split architecture enables a diverse ecosystem of engineers and domain experts to focus on their area of expertise. 
For instance, by default, a human reviewer need not worry about feature engineering or model training.
Similarly, an engineer deploying models need not focus on feature generation or data curation. 
While more ``end-to-end" methods for cross-modal adaptation exist (see Section~\ref{sec:relwork}), we find this split architecture effectively trades-off between generality and ease of maintenance and deployment. 

\subsubsection{Pipeline Steps}

Given that we preserve the traditional \splitarch, we augment each step for a cross-modal setting as follows:

\minihead{Feature Generation (Figure~\ref{fig:system}A, \S\ref{subsec:feat_gen})}
We generate features common to both new and existing data modalities.
To do so, we identify and apply \orgress that process data points of multiple modalities, and output values in a space common between them (as in Figure~\ref{fig:feature_gen}). 
These features lay the foundation to using data across modalities in the subsequent pipeline steps.
This step addresses the production challenge of processing rich data modalities, and the cross-modal challenge of bridging the modality gap.

\minihead{Training Data Curation (Figure~\ref{fig:system}B, \S\ref{subsec:data_curation})}
We automatically generate labels for the new, unlabeled data modality to develop a training dataset. 
To do so, we perform weak supervision~\cite{snorkel} using new methods for automatic labeling function creation via frequent itemset mining~\cite{fim} and label propagation~\cite{labelprop} that leverage the shared features from the first step. 
This step addresses the production challenge of leveraging diverse information sources, and the cross-modal challenge of labeling rich modalities.

\minihead{Model Training (Figure~\ref{fig:system}C, \S\ref{subsec:cross_modal_arch})}
We train a model using both the weakly supervised data in the new modality and fully supervised data of existing modalities. 
To do so, we combine the features generated in the first step to construct a vector feature representation by concatenating the generated common features directly.
We evaluate three approaches, but find that simple concatenation outperforms alternatives. 
This step addresses the cross-modal challenge of combining all available resources at deployment. 
\section{Feature Generation}
\label{subsec:feat_gen}

In this section, we describe the first step of our cross-modal pipeline.
As input, we are given data of a new modality (e.g., video posts), and must train models for existing tasks.
We have labeled data and models that perform these tasks for existing data modalities, but cannot be directly used due to the modality gap (Challenge 2 in \S\ref{subsec:cm_challenges}).
Further, processing new, rich data modalities is time and resource intensive (Challenge 3 in \S\ref{subsec:design_considerations}).
We describe how to overcome these hurdles by developing structured (i.e., categorical and quantitative) features common across data modalities via \orgress (see Figures~\ref{fig:feature_gen} and~\ref{fig:system}A).

\subsection{Generating a Common Feature Space} As ML becomes more common across domains, \orgress \eat{in the form of services, models, statistics and heuristics} are being curated by industrial product teams and labs~\cite{OR_googlecloud,OR_aws,OR_pinterest}, stand-alone companies~\cite{OR_quandl,OR_onnx}, and academic research groups~\cite{OR_stanfordnlp}. 
These resources take as input data points of various modalities, and return categorical and quantitative outputs in the form of features, class metadata and statistics that describe these data points. 
As a result, while a user has yet to develop models for their specific tasks in the new modality, they can still apply these \orgress to data points of the new modality.

We directly apply \orgress to transform new and existing data modalities to a common feature space (see Figure~\ref{fig:system}A).
The output of each resource corresponds to a numeric feature or multivalent categorical feature (equivalently, a set of one-hot-encoded features). 
A set of $k$ resources will return $k$ features, $f_i$ for $i \in \{1,...,k\}$.
We denote the resulting common feature space as $F = \{f_1,...,f_k\}$, with a data point $x$ being represented as $F_x = \{f_{1}(x),...,f_{k}(x)\}$.
As an example, consider a text and image post represented as data points $t$ and $i$, respectively.
We have services to detect the presence of profanity in text ($T_{\text{profanity}}$) and image posts ($I_{\text{profanity}}$), and to detect their setting ($T_{\text{setting}}, I_{\text{setting}}$).
Applying these resources to each data point provides us with the following feature representation ($F_t, F_i$): 

{\small $$F_t = (T_{\text{profanity}}(t) , T_{\text{setting}}(t)) = (f_{1}(t), f_{2}(t)) = (True, outdoor),$$
$$F_i = (I_{\text{profanity}}(i), I_{\text{setting}}(i)) = (f_{1}(i), f_{2}(i)) =  (False, outdoor).$$}

\eat{We find these features sufficient for our tasks, but note that it is the first step in developing more complex embedding-based feature spaces~\cite{huse,emb2}.}

In this work, users must curate applicable \orgress. This may be challenging for proprietary or less-supported modalities (e.g., graph-structured data), and we discuss how developing systems for \orgress discovery can help in Section~\ref{subsec:future_feat}.

\subsubsection{Examples of \OrgRess}
\label{subsec:or_ex}
\eat{We categorize common classes of organizational resources that enable transformations as follows:}

\minihead{Model-Based Services} 
Teams can access classification and processing services that operate over new and existing data modalities. 
Examples include: topic models that categorize content; motif discovery tools to transform time series to categorical patterns; knowledge graph querying tools to extract entities and relationships from data points.
\vspace{3pt}

\emph{In our moderation example, the organization has topic modeling services that map text or image and data points to a common set of categories (e.g., brands), services that caption images, and tools that split a video into representative image frames.
To featurize data points for the new video data modality, the team can extract frames from a video post to create image data points using the video splitting tool.
The extracted image data points can then be used as input to the topic modeling and captioning services to generate a shared features between video, image, and text posts.}

\minihead{Aggregate Statistics and Metadata} 
Teams collect statistics fitting their applications' and customers' needs.
Organizations also possess metadata to track data points across different teams (e.g., user or post ID), enabling these statistics to be used as features across modalities.
\vspace{3pt}

\emph{In our moderation example, a team that identifies problematic users may track the number of times a user is reported, and a content recommendation team might track how many times a post is shared. 
The content moderation team can use the user ID metadata field to connect moderated posts with the statistics from these two teams.
They can then use these statistics as input features in determining if a post may be policy-violating (e.g., a post from a user who is often reported and posts policy-violating content that spreads quickly is more likely to continue violate policies).}

\minihead{Rule-Based Services} 
Teams develop heuristics and rules to make manually collecting, analyzing and labeling data more efficient. 
For instance, to sample candidate positive examples in class-imbalanced scenarios, experts first use heuristics to justify transitioning to automated methods including active learning~\cite{activelearning}.
\vspace{3pt}

\emph{In our moderation example, the team may know that certain keywords are related to sensitive content, or that certain user behaviors are correlated with spammers. 
They may have used these rules to sample candidate training data to review for sensitive content, and can use them as binary features.}
\eat{(e.g., creating a new account and adding several unrelated users as friends)}

\section{Training Data Curation}
\label{subsec:data_curation}
Once we generate common features across data modalities, we must curate labeled examples for model training.
As we show in Section~\ref{subsec:eval_modality}, using the shared features to train a model with the labeled data of existing modalities performs worse than training a model on the target modality.
We instead demonstrate how to leverage existing modalities to generate labeled data in the target modality \emph{without} additional human labeling (Challenge 1 in \S\ref{subsec:cm_challenges}).
We achieve this via weak supervision (WS), which also allows us to use features unavailable at deployment time (Challenge 4 in \S\ref{subsec:design_considerations}). 
We now introduce WS, and describe how to use our common feature space to overcome three challenges in using WS for cross-modal adaptation (see Figure~\ref{fig:system}B).

\subsection{Introduction to Weak Supervision}
WS uses cheap but noisy labels to curate training data.
Snorkel is a WS framework where users generate \emph{labeling functions} (LFs) to programmatically label groups of data points~\cite{snorkel}.
To label a set of unlabeled data points, $\mathcal{X}$, Snorkel's pipeline proceeds as follows:
\begin{enumerate}
    \itemsep-.2pt
    \item \textbf{Develop LFs.} Domain experts use a small, labeled development dataset to create LFs. LFs are functions that take a data point and all related features as input, and output a label or abstain (e.g., in a binary setting, an LF returns positive, negative, or abstain). In our moderation example, a sample LF may be: \emph{if a post contains excessive profanity it is harmful speech, else abstain.} Snorkel requires both high precision and high recall LFs that each perform better than random.
    \item \textbf{Programatically apply generated LFs to $\mathcal{X}$.} Unlike a standard labeling pipeline, $\mathcal{X}$ can be very large as labels are not human-generated. Snorkel performs best when many data points in $\mathcal{X}$ have LFs that return labels instead of abstaining, i.e., have high coverage.
    \item \minihead{Learn probabilistic labels from Step 2} Snorkel uses a generative model to estimate each LF's accuracy by evaluating correlations between them when applied to $\mathcal{X}$. The estimated accuracies are used to return a weighted combination of the weak labels applied to each data point, i.e., probabilistic labels.
\end{enumerate}

The probabilistically labeled data can be used to train an end discriminative model for the target task that operates over a noise-aware loss function.

This entire process is offline---the LFs used in training are only used to generate training data, and are not used when models are served. As a result, we can generate probabilistic training data using features that are not feasible to compute or obtain at serving time (i.e., nonservable features). 
\revision{We leverage Snorkel as one of many primitives in our end-to-end pipeline.
While Snorkel supports both binary and multi-class classification tasks, in this work, we evaluate our methods on binary classification tasks, but can easily extend to multi-class.
Snorkel does not apply to regression tasks. 
However, the feature space we induce via \orgress can be used for tasks including similarity search and clustering, as we use for label propagation in Section~\ref{subsec:lp}. 
}

We build on Snorkel Drybell's industrial case study~\cite{drybell} to extend WS for cross-modal adaptation.
In this process, we encountered three challenges in developing LFs at scale.
We now describe how to use \orgress to overcome each (see Figure~\ref{fig:system}B).

\subsection{Curating Structured Data and a Dev Set}
Users typically construct LFs by defining predicates that operate over their data points.
This is challenging in our setting for two reasons.
First, generating LFs over unstructured data modalities (such as image and video) is an active area of research~\cite{coral, dugong, rekall, sequential1,sequential2}.
In contrast, defining predicates is straightforward over structured feature spaces such as text (e.g., string or pattern matchers), quantitative data (e.g., thresholds), or categorical data (e.g., checking presence of a topic) variables.
For instance, lacking captions or metadata, existing work to detect if an image post displays a sporting event requires off-the-shelf classifiers to identify the setting, players and spectators, and rules to verify that they coexist in the right locations. 
This is difficult for nuanced tasks, such as sensitive content identification.
Second, users require a labeled development set to develop and validate the performance of candidate LFs.
In a cross-modal setting, neither of these data nor label requirements are met.

We overcome these challenges by leveraging the common feature space induced in our first pipeline step. This feature space provides categorical outputs (e.g., post setting or objects) over which users can define LFs for both new and existing data modalities. 
This feature space also enables users to leverage labeled data of \emph{existing} data modalities as a development set for LF creation in the new modality.
Users can thus define LFs over features common to existing and new modalities, and evaluate LF performance using the labeled data from existing modalities. 
For instance, in the moderation example, if an expert knows that specific topics in text or image are frequently flagged as spam, then splitting a video into image frames and running topic models on the images would enable the same LFs to apply to video.

\subsection{Creating LFs Without Domain Expertise}
In a typical WS deployment, domain experts construct LFs using their expertise.
They must understand how classification tasks vary across language, country, and region, as well as how the task evolves in a new data modality.
However, not all teams have access to domain experts who can swiftly develop LFs for their tasks in the new data modality.
Prior work in automatic LF generation can overcome this challenge, including model-based approaches such as Snuba~\cite{snuba}.
We found such methods difficult to immediately integrate (and justify) with existing production workflows and infrastructure given limited developer capacity.
In response, we developed a method based on frequent itemset mining to automatically generate LFs that results in less engineering overhead and outperformed our experts.

Our method mimics domain-experts via rules that are a conjunction of feature values. 
To construct an LF, we identify feature combinations that occur more often in positive than negative examples, and vice versa. 
We first select feature values that---when used as an LF---meet pre-specified precision and recall thresholds over the development set.
Higher order feature combinations are added when they meet the threshold, as in the Apriori algorithm~\cite{fim}.
In our experiments, we found order-1 sufficient in practice.

To minimize correlations across LFs, each LF is a conjunction of feature values identified by the mining procedure, defined over a single feature. 
To decrease runtime in class-imbalanced scenarios, similar to difference detection and explanation in large scale data~\cite{macrobase}, we first mine for candidate feature values in the positive examples.

We evaluate the gap in performance (in terms of precision, recall, and F1 score) and development time between our automatically- and expert-generated functions for one classification task in Section~\ref{subsec:eval_ws_human}.
We find our approach straightforward for an engineer to deploy and maintain as opposed to leveraging complex techniques that, for instance, require evaluating and maintaining model ensembles.

\revision{
While Snorkel enables users to automatically construct labeling function generators~\cite{snorkel}, these methods still require domain knowledge to align data sources and dictate how to construct LFs (e.g., \emph{knowledge graph relationships X are sensitive}). Our method alleviates these needs if a non-expert engineer is developing pipelines as a prototype or finalized product.
Validating our approach's performance provides justification to invest resources to develop custom interfaces and pipelines for our domain experts similar to~\cite{snorkel}, as may be needed for more nuanced future tasks (see Section~\ref{subsec:future_labeling}).}

\subsection{Finding Borderline Examples}
\label{subsec:lp}
WS requires positive and negative LFs with high precision, recall, and coverage.
We found that users struggle to create LFs with high recall and coverage, and LFs that capture the negative class' behavior.
This occurs as it is straightforward to develop rules that describe ``easy" examples in the positive class when basic modes are well defined (e.g., profanity to identify hate speech).
In contrast, the behavior of borderline positives and the negative class is vast and unspecified---especially in class imbalanced settings.
In response, we leverage our common features to identify and propagate labels to data points in the new modality that are similar to labeled examples in previous data modalities via \emph{label propagation}~\cite{labelprop} (as is common in active learning).

To perform label propagation, we must determine what labels to propagate, and how to propagate them. 
For the former, we can propagate weak labels (LF outputs), probabilistic labels (Snorkel outputs), or human-generated labels. 
While each of these label sources demonstrated promise in early experimentation, we use the third as they would be least impacted by biases and inaccuracies in our LF generation method. 
For the latter, we first develop a \revision{graph $\mathcal{G}$} induced by our common feature representation (Figure~\ref{fig:system}B).
An unlabeled data point that shares edges with labeled data points is assigned a weighted combination of its neighbors' labels.
We update this assignment to convergence to return a weighted score.
This score is used to construct a threshold-based LF, but can also be used as a form of probabilistic label.
We again leverage a development set of labeled examples in existing modalities to tune this threshold.

To construct $\mathcal{G}$, each data point (across all modalities) corresponds to a vertex $V_i$. 
We construct edges between vertices $i$ and $j$ with weight $w_{ij}$ assigned based on the similarities between their feature representations $F_i, F_j$---their Jaccard similarity if the features are categorical, or a pre-specified distance metric if the features are quantitative or based on pre-trained embeddings (see Algorithm~\ref{algo:graph_build}).
As an example, an edge between text data point $t$ with $F_t = (True, outdoor)$ and image data point $i$ with $F_i = (False, outdoor)$ returns $w_{ti} = 1$. 
In practice, each feature's contribution is normalized in lines 5 and 7, which we omit for simplicity.

\begin{algorithm}[t!]
\revision{
\begin{algorithmic}[1]
\small
\Statex \textbf{Input:}  $F_i, F_j$: data points' features
\Statex Let $F$ denote the set of all features instantiated by $F_i, F_j$
\Statex \textbf{Output:} $w_{ij}$: weight between input data points 
\Statex \hrule
\Function{compute-weight}{$F_i, F_j, F$}:
	\State Initialize: $w_{ij} = 0$
	\For{$k = 1,...,|F|$}  
		\If{$f_k$ is numeric} 
		\State $w_{ij} = w_{ij} + \|f_{k}(i) - f_{k}(j)\|$
		\Comment{any norm, e.g. $\ell^2$}
		\EndIf 
		\If{$f_k$ is categorical} 
		\State $w_{ij} = w_{ij} + \text{jaccard}(f_{k}(i), f_{k}(j))$
		\EndIf		
	\EndFor
	\\\Return{$w_{ij}$}
\EndFunction
\end{algorithmic}
}
\caption{Graph Weight Computation}
\label{algo:graph_build}
\end{algorithm}

Label propagation can leverage features that are difficult to construct LFs with as long as a distance metric can be defined for them; we use features specific to the new modality to construct edges, including unstructured features such as image embeddings.
As label propagation is too costly to run at deployment time, we leverage this as nonservable information for training data curation to boost end-model performance. 
We evaluate its effectiveness in Section~\ref{subsec:eval_lp}.

\begin{figure}
  \includegraphics[width=\linewidth]{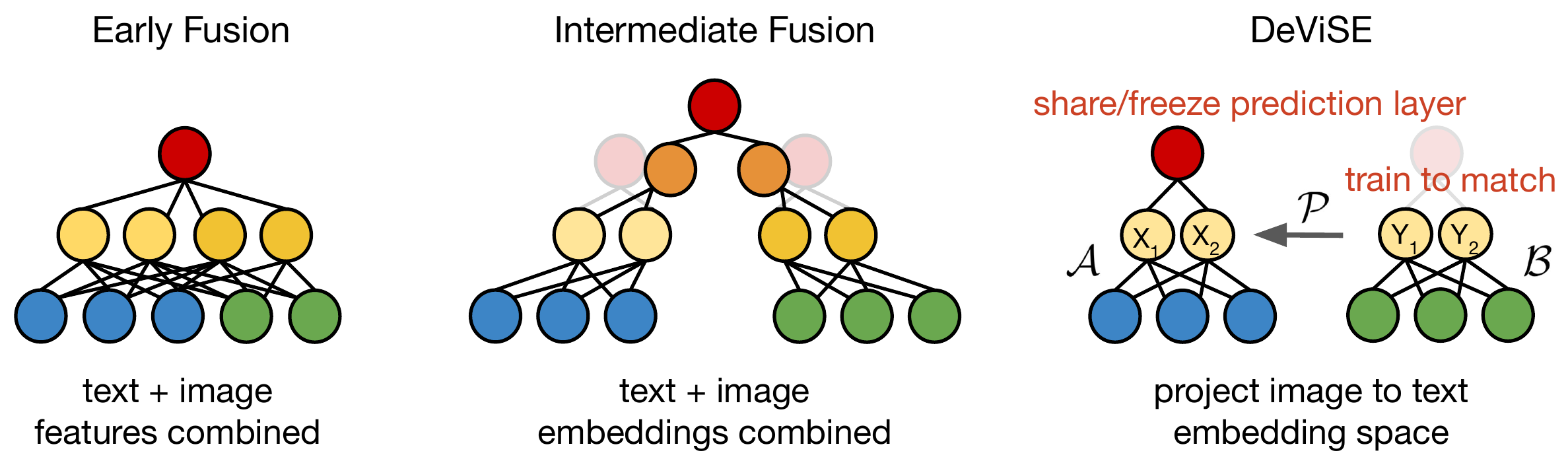}
  \caption{Three means of performing cross-modal training, displayed in the two-modality setting.}
  \label{fig:model_train}
\end{figure}

\section{Model Training}
\label{subsec:cross_modal_arch}
Following the curation of labeled training data, we must train a model for classification. 
Our goal is to leverage information across all of the generated modalities and label sources to train a better model for the new data modality (Challenge 3 in \S\ref{subsec:cm_challenges}).
\revision{Prior work in cross-modal training assumes one-to-one mappings between modalities~\cite{crossmodalws,mml1,mml2}, and that operating directly over the new modality provides high-quality features~\cite{mmdbms1}. However, we find that operating directly over our feature space performs up to 1.54$\times$ better than operating directly over the new modality (see Section~\ref{subsec:eval_modality}). Thus, we evaluate} three methods to jointly train over new and existing modalities and label sources under our induced feature representation (Figure~\ref{fig:model_train}):

\minihead{Early Fusion}
We merge the features of all data modalities to create a single common representation as in Figure~\ref{fig:feature_gen}.
Features shared by all data modalities are merged into a single field (e.g., raw text from text posts, and captions derived from image data points), and features specific to certain data modalities are left empty for those that do not have these features present (e.g., image-specific embeddings will not be present in text data).
All data modalities and label sources are then trained \revision{by combining them into a single dataset}. 


\minihead{Intermediate Fusion}
We learn an embedding for each data modality, and concatenate these embeddings as input to a final, jointly trained model. 
Training proceeds in two passes over the training data.
In the first stage, we create independent models for each data modality.
In the second stage, we remove the final prediction layer (e.g., softmax) from each of these models. 
We then perform a second pass over all of the data, where the shared features are passed into \emph{all models} in which they exist.
The model outputs are concatenated to create a new feature embedding. 
This embedding is used as input to a final model for training. 
\new{Our motivation in constructing this architecture is that data modalities with fewer data points may get overpowered in the early fusion model. 
By training each modality independently prior to concatenation, we hope to alleviate this.
}

\minihead{DeViSE~\cite{devise}}
We learn an embedding using existing data modalities, and then project data points from the new modality to the embedding space for classification.
\revision{The original DeViSE algorithm is a classic cross-modal baseline that we have adapted to our setting} as follows (see Figure~\ref{fig:model_train}).
First, we train a model $\mathcal{A}$ over existing data modalities as in early fusion, similar to DeViSE's language model pre-training.
This model is then ``frozen," so none of its parameters change. 
Next, we pre-train a model $\mathcal{B}$ over the weakly supervised data of the new modality, similar to DeViSE's visual model pre-training.
In the final training stage, we pass points of the new modality to $\mathcal{B}$ and simultaneously pass the shared features between the existing and new modalities as input to $\mathcal{A}$, and compute the model outputs prior to the final prediction (e.g., softmax) layer of both $\mathcal{A}$ and $\mathcal{B}$, which we denote $X$ and $Y$, respectively. 
We then train a ``projection layer" $\mathcal{P}$ (as in DeViSE) to match $Y$ with $X$.
At inference time, we pass incoming data points  through $\mathcal{B}$ and the projection layer $\mathcal{P}$, and use the final prediction layer of the initially trained, frozen $\mathcal{A}$.

\begin{table}[]
\begin{center}
\caption{We report the number of labeled text data points, unlabeled image data points that we label via weak supervision, labeled image data points used as a test set, and the test set positive rate for each task.}
\label{table:datasets}
\begin{tabular}{lrrrr}
\hline
Task            & $n_{\text{lbd,text}}$ & $n_{\text{unlbld,image}}$     & $n_{\text{{lbd,image}}}$  & \% Pos \\ \hline
CT 1   & 18M                 & 7.2M                          & 17k                       & 4.1\%     \\
CT 2   & 26M                   & 7.4M                          & 203k                      & 9.3\%     \\
CT 3  & 19M                   & 7.4M                          & 201k                      & 3.2\%     \\
CT 4   & 25M                   & 7.3M                          & 139k                      & 0.9\%     \\ 
CT 5   & 25M                   & 7.4M                          & 203k                      & 6.9\%        \\ \hline 
\end{tabular}
\end{center}
\end{table}

\section{Evaluation: Google Case Study}
\label{subsec:case}

We present our case study in cross-modal adaptation at Google.
We describe our \ch{5} classification tasks, \orgress and experimental setup, and show that:

\begin{enumerate}
    \item \textbf{End-to-End (\S\ref{subsec:eval_e2e}): } Our cross-modal pipeline outperforms a fully supervised pipeline that uses up to \ch{750k} data points with respect to AUPRC.
    \item \textbf{\OrgRess (\S\ref{subsec:eval_orgress}): } Model performance with respect to AUPRC scales with the amount of \orgress used.
    \item \textbf{Modalities (\S\ref{subsec:eval_modality}): } Training a model using human-labeled data of existing data modalities is up to \ch{1.63$\times$} less effective than using weakly supervised data of the target modality with respect to AUPRC.
    \item \textbf{Modalities (\S\ref{subsec:eval_modality}): } Jointly training data modalities is up to \ch{1.63$\times$} and  \ch{1.23$\times$} better than training on text or image in isolation, respectively, in terms of AUPRC.
    \item \textbf{WS (\S\ref{subsec:eval_ws_human}): } Automatic LF creation is up to 1.87$\times$ faster than manual development by a human expert, which required 7 hours spread over 2 weeks, and performs better by 2.71 F1 points. 
    \item \textbf{WS (\S\ref{subsec:eval_lp}): } Label propagation  complements our high-precision LFs to improve recall by up to 162$\times$, providing up to \ch{1.25$\times$} improvement in F1 score. 

\end{enumerate}

\subsection{Classification Tasks}
We evaluate \ch{5} binary topic or object classification tasks developed by an engineering team (see \revision{Table~\ref{table:datasets}}). 
\revision{In topic classification, engineers classify if a given entity (e.g., user post) represents a topic of interest (e.g., explicit content or hate speech).
In object classification, the engineers classify if a given entity contains a specific object or object type (e.g., illegal products).} 
We consider a two-modality setting where models are trained for text entities and must now apply to image entities.
T refers to a fully supervised text model, and I refers to a weakly supervised image model.

We treat a modality with curated, human-labeled data (i.e., image) as the ``new" modality. 
\revision{Labeled data is sampled uniformly at random over all curated data points prior to a specified point in time. 
We sample live traffic after this time to generate unlabeled data independent of previously labeled image data, ensuring no train-test leakage.} 
\eat{By definition, labeled data in the new modality does not exist in a cross-modal adaptation setting, and we have large volumes of unlabeled live data.} 

Figures~\ref{fig:rc_cross_over}-\ref{fig:rc_sources} reflect the results from CT 1. 
\revision{We focus on CT 1 for these feature-related microbenchmarks as it captured a majority of the different possibilities we saw when adding new features (substantial gain, little gain, and no gain), but provide high-level details for the other tasks in the text}, with all end-to-end results in \revision{Table~\ref{table:results}}.

\subsection{Team-Specific \OrgRess}
We leverage \orgress across two of the categories described in Section~\ref{subsec:feat_gen}:

\minihead{Model-Based Services}
Google has several custom topic models, knowledge graphs, named entity recognition models, and object detection models that are maintained for use across the organization.
Teams query these services to generate information such as topic hierarchies and categorizations, and language translations.
\revision{We use two types of model-based services: topic-model-based, and page-content-based.
The former refers to topic-models applied directly to the data points.
The latter refers to models that apply to web pages and auxiliary information regarding the data points.
Each provides different views into a data point, and contributes differently to each task. 
For instance, in the social media example, policy violations may be defined in terms of a user's post or content that a user’s post links to. 
}

\minihead{Aggregate Statistics and Metadata}
Teams can extract metadata including user ID, customer ID, URL, keywords, and categorization from a data point.
The engineering team has been deploying classification models while collecting this metadata for several years.
As a result, they can compute aggregate statistics from the outputs of these models across users, customers, URLs, topics and categories. 
\revision{
We use keyword-based and URL-based metadata services.
}

We use 15 services \revision{to generate 15 features}: 14 are categorical and multivalent \revision{with vocabularies of up to several thousand categories}, and two are nonservable (including the output of label propagation as in \S\ref{subsec:data_curation}). 
In addition, images possess 3 pre-trained embedding and image-specific \revision{features}.
\revision{We evaluate four types of services used to generate feature sets: URL-based, keyword-based, topic-model-based, page-content-based, labeled as} sets A, B, C, and D, which provide us with 3, 2, 5, and 5 features, respectively. 
We state which features are included in the discriminative model for each modality as T + [ABCD]* and I + [ABCD]*. 

\revision{
Services we use are pre-computed for each data point as the generated features assist teams across the organization. 
Thus, we leverage these features without incurring additional overhead to generate them.
When an organization curates resources from scratch, they must account for the potentially large computational overheads that are incurred from applying \orgress to new modalities for feature generation---especially for large-scale datasets in rich modalities such as video, which require heavy processing. 
}

\subsection{Experimental Setup}

\minihead{Implementation} 
\new{We implement the feature engineering and LF pipeline using our MapReduce framework.}
\new{We use Snorkel Drybell~\cite{drybell} as a scalable WS system, and Expander~\cite{expander}, a large-scale graph-based machine learning platform, for streaming, distributed label propagation~\cite{expanderalgo}.}
\new{For model training and serving, we use TFX~\cite{tfx}.}
We leverage the same models used by the engineering team in our case study, modified to train with probabilistic labels using a cross-entropy loss function,  with hyperparameters set by Vizier~\cite{vizier}.
For multi-modal training, this refers to the models for each modality prior to embedding concatenation.
The models supported by the team's TFX pipelines are logistic regression and fully-connected deep neural networks, where the best performing is used in production. 
\new{We report performance over the neural network models for CT 1-4, and logistic regression for CT 5 due to improved performance.}

\begin{figure}
  \includegraphics[width=\linewidth]{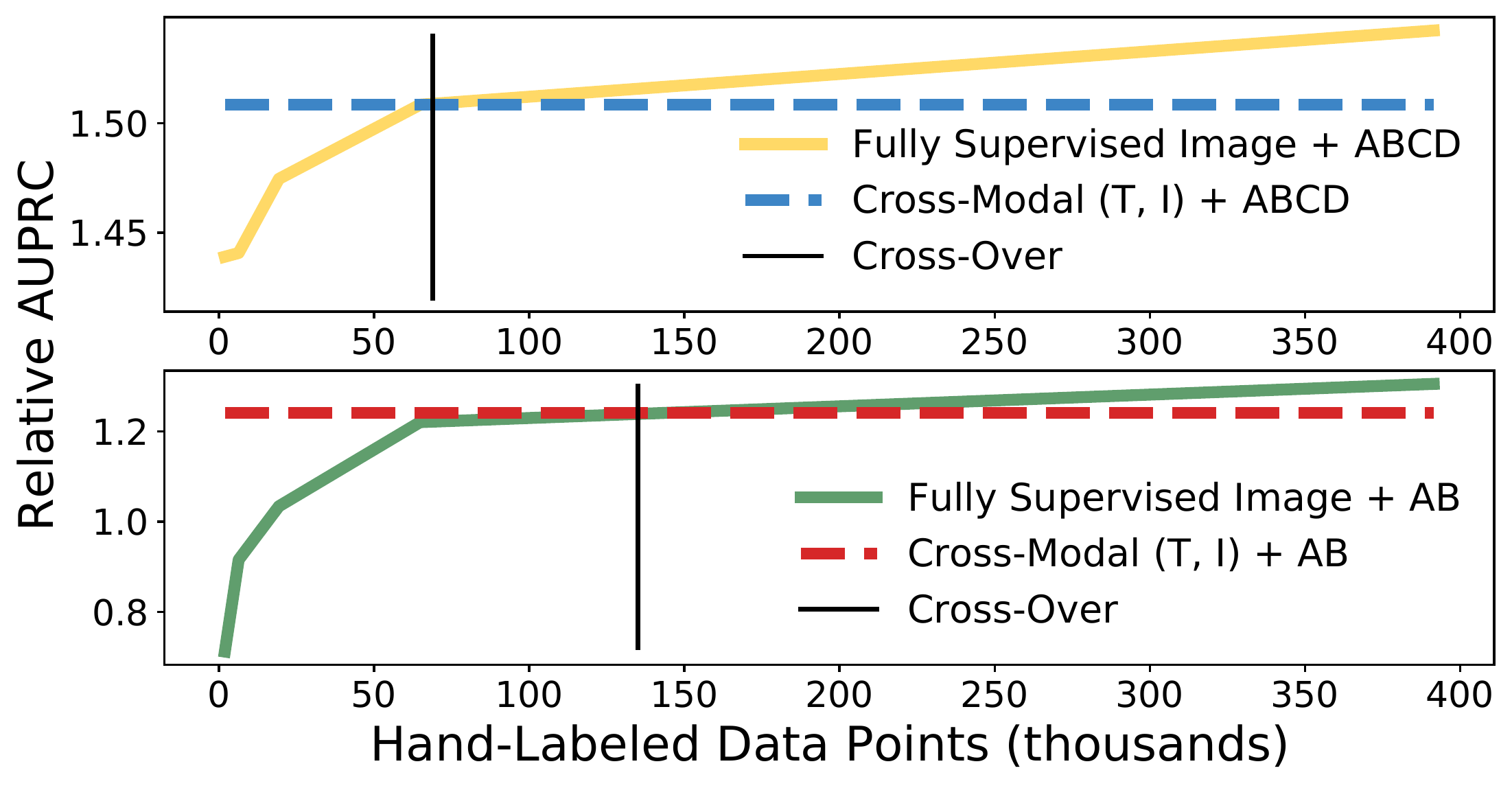}
  \caption{AUPRC of our cross-modal model, and a fully supervised image model with two or four sets of features, relative to a fully supervised model trained on pre-trained image embeddings for CT 1. We require around 60k hand-labeled images to outperform our approach when using all four sets (top), and around 140k when using two sets (bottom).}
  \label{fig:rc_cross_over}
\end{figure}
\begin{figure*}
  \includegraphics[width=\linewidth]{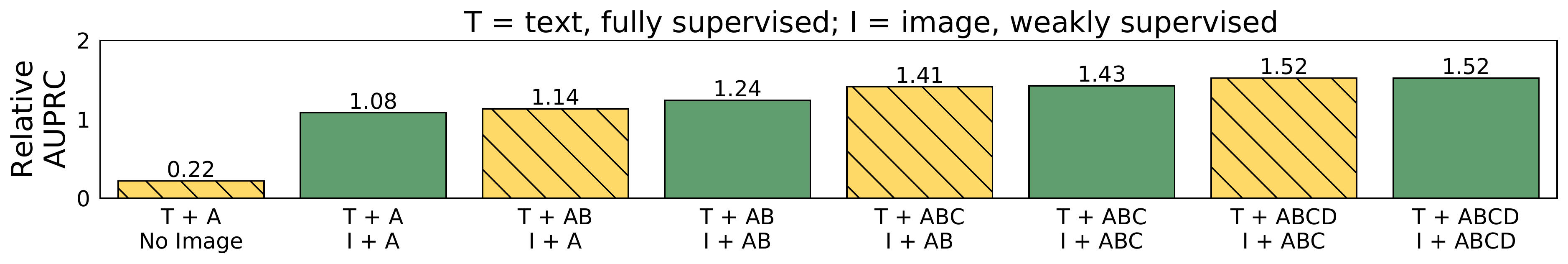}
  \caption{Factor analysis for CT 1 demonstrating the increase in AUPRC when adding additional sets of features (A,B,C,D) for each data source (T, I), relative to a baseline fully supervised model trained using pre-trained image embeddings.
  We show that for this task, adding new features and data improves AUPRC. After including image data (first to second), we find adding features (solid to hashed) improves performance more on average than adding modalities with that feature (0.11 vs 0.04).} 
  \label{fig:rc_lesion}
\end{figure*}
 
\minihead{Evaluation Metric}
We compute the area under the precision-recall curve (AUPRC) over the labeled image test set to evaluate our pipeline. 
\new{Depending on the task sensitivity in the organization, the cut-off to compute metrics including F1 score are decided upon viewing live performance, rendering AUPRC more representative for offline evaluation.}
We report AUPRC relative to a baseline fully supervised image model trained with only pre-trained embedding features.

\subsection{End-to-End Comparison}
\label{subsec:eval_e2e}
In Figure~\ref{fig:rc_cross_over}, we compare the performance of our pipeline for cross-modal adaptation with a hand-labeled, fully supervised model using features from all services (top) and with two sets of services (bottom) for CT 1.
We use all \revision{features} to generate LFs for weak supervision in both cases. 

We demonstrate two takeaways in this experiment. 
First, we show that our architecture can save weeks to months of human-curation time, depending on how many resources are expended for the task.
We see that the cross-over point at which a fully supervised model outperforms our cross-modal pipeline: 60k (top) or 140k (bottom) data points, representing substantial human labeling effort. 
Second, we show that using nonservable features~\cite{snorkel} to develop LFs improves the performance of the weakly supervised cross-modal pipeline relative to a fully supervised baseline without these features (bottom vs. top).
\revision{We include fewer features for the discriminative end-model than the LFs to mimic nonservable features in Figure~\ref{fig:rc_cross_over} (bottom). We see that using feature sets C and D only for weak supervision (i.e., via LFs) and not the fully supervised model (bottom) requires more hand-labeled data points to meet the performance of the weakly supervised model (140k vs. 60k). }

For the remainder of the tasks, we display the cross-over point at which a fully supervised model outperforms our cross-modal pipeline in \revision{Table~\ref{table:results}}.
\revision{Similarly, the relationship between the servable and simulated ``nonservable" use case holds for these other tasks with the difference ranging from 400 to 190k data points.}
Regardless of the exact cross-over point, our cross-modal pipeline enables us to deploy models in production without waiting for domain experts to be trained for the task.
\revision{For our non-mission-critical tasks (e.g., internal or non-user facing tasks) with safeguards in place (either in the form of heuristics, or human reviewers with understanding of the task for alternative modalities), this enables rapid initial model deployment that can be augmented via techniques for active learning~\cite{activelearning} or self-training~\cite{self} on the order of days. 
However, in the usual case, where a \emph{representative, unbiased} test set is required to validate a model for production, the model deployment time will be on the order of weeks, as our methods only reduce, not remove, the time needed to curate data for both training and validation.}

Table~\ref{table:results} demonstrates a wide range of cross-over points across each task.
\revision{We believe this range is a result of the relative difficulty in modeling each task with our manually curated features.
In tasks where our features provide adequate discriminative capabilities, our labeling functions can capture the behavioral modes of positive and negative examples. For instance, in the social media example, detecting politically-inclined posts (i.e., those with candidate faces, or party imagery) may be easier than detecting false news. As we verify for CT 1 in Section~\ref{subsec:eval_ws_human}, tasks that perform well (CT 1, 2, 6) likely exhibit this behavior.}
Understanding which in regime we are operating is currently a manual procedure, and an area for future work (see Section~\ref{subsec:operating_regime}). 

\eat{In addition, in the case of business-critical tasks that require human reviewer approval, we can use cross-modal models for active learning~\cite{activelearning} to assist with sampling points for collecting human-labeled ground-truth.}

\begin{table}[]
\begin{center}
\caption{AUPRC for a fully-supervised text model (T + ABCD), a weakly-supervised image model (I + ABCD) and a cross-modal model (T, I + ABCD) using all four sets of \revision{features}, relative to a fully supervised image model trained with only pre-trained image embedding features. We report the number of fully-supervised image examples required to outperform our approach (i.e. ``cross-over" point).}
\label{table:results}
\begin{tabular}{lrrrr}
\hline
Task                & Text          & Image         & Cross-Modal   &  Cross-Over          \\ \hline
CT 1        & 1.12          & 1.43          & 1.52          &   60k examples    \\
CT 2       & 1.49          & 2.32          & 2.43          &   50k examples     \\
CT 3       & 0.88          & 0.95          & \ch{1.14}          &   5k examples     \\ 
CT 4       & 1.74             & \ch{2.00}             & 2.45          &   4k examples     \\ 
CT 5       & 1.67       & 2.03       & 2.42       &   750k examples     \\ \hline
\end{tabular}
\end{center}
\end{table}

\subsection{\OrgRess Factor Analysis}
\label{subsec:eval_orgress}

In Figure~\ref{fig:rc_lesion}, we perform a factor analysis to show that adding features and data modalities (i.e., \orgress) incrementally improves end-model performance for CT 1. 
At each step of the factor analysis, we add a new feature to either the text modality or the image modality. 
We train an early fusion model as described in Section~\ref{subsec:cross_modal_arch}. 
We typically observe that the addition of new features improved performance more significantly than the inclusion of more data modalities containing these features (in Figure~\ref{fig:rc_lesion}, these are steps from solid to hatched bars).
We note that this behavior can be dependent on the task and the relative distribution differences across features in different modalities. 
\revision{For instance, we found no improvement from adding image features D (final hatched to solid increment) for CT 1, and little improvement from adding image features C. 
We found these results to hold true across our tasks, with relative AUPRC improvements of 0 to 6.23$\times$. 
This may have occurred as our feature space was constructed with manually curated \orgress that resulted in similar feature distributions across text and image data points. 
However, a low quality feature/\orgres might negatively impact performance if it were selected via automated processes without validation; not all features may be relevant for the downstream task, thus quality must be validated in advance. 
Identifying how to best weight and value candidate \orgress is critical to scaling our proposed techniques. We are evaluating scalable methods for domain adaptation~\cite{frustratinglyeasydomainadaptation, domainadaptation1,crosstrainer}, which together with feature attribution~\cite{feature1,feature2,feature3} indicate promise in understanding which features to leverage from each modality.}

\subsection{Multi-Modal Training Lesion Study}
\label{subsec:eval_modality}
\begin{figure}
  \includegraphics[width=\linewidth]{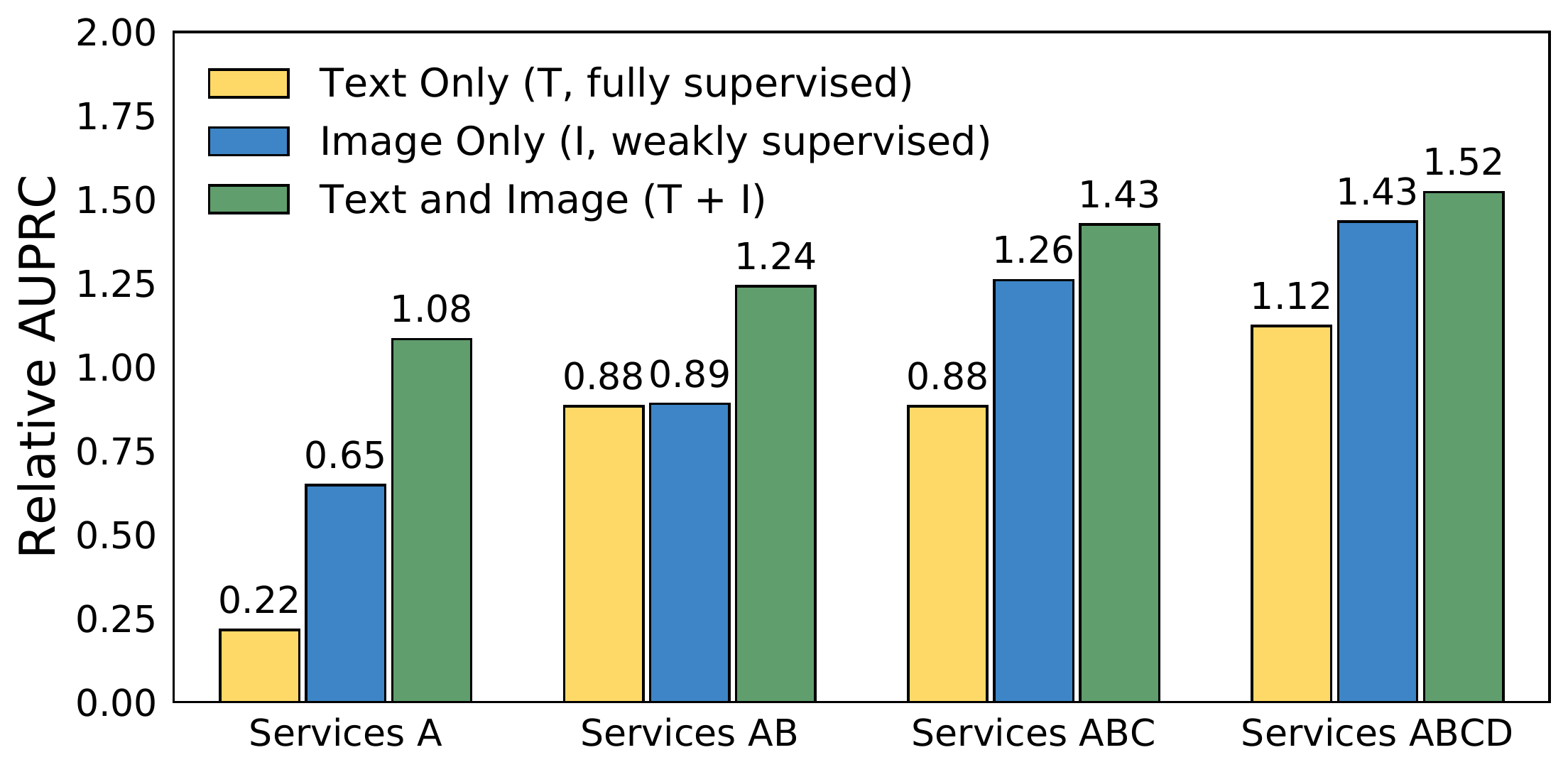}
  \caption{AUPRC of text, image, and cross-modal models, relative to a fully supervised image model trained with just pre-trained image embeddings for CT 1. Combining modalities is better for each feature space induced by the services (A,B,C,D).}
  \label{fig:rc_sources}
\end{figure}
In Figure~\ref{fig:rc_sources}, we perform a lesion study over data modality for CT 1 as we augment our feature set.
We first find that training a fully supervised model over existing data modalities (i.e., text) and using it for inference on a new target modality (i.e., image) is not as effective as a weakly supervised model trained in the target modality.
A model trained on 7.2M weakly supervised images (Figure~\ref{fig:rc_sources}, blue) outperforms one trained on a fully supervised text dataset of 18.4M examples (Figure~\ref{fig:rc_sources}, yellow) despite lacking human-curated labels.
\revision{The exact improvement is dependent on the features used. For instance, the difference when using two features (A, B) is insubstantial.  
Table~\ref{table:results} shows this result holds for all other tasks, with improvements ranging from 1.1-1.6$\times$ in the case with all features used.}
Using fully supervised data of the target modality only increases the performance gap: in CT 1, the text model performed 1.4$\times$ worse despite having 48$\times$ more data.
\revision{This result holds true across all tasks, and, as seen in Table~\ref{table:results}, CT 3 and CT 4 in particular, which have much lower cross-over points.}
We hypothesize this occurs as despite the common feature space, the input distribution is not identical across modalities.

Second, we find that combining data modalities (Figure~\ref{fig:rc_sources}, green) improves performance compared to using any one modality.
This is true as we augment our feature set, \revision{and for CT 2 - 5.}
We present the results for the case with all features A,B,C,D for other tasks in \revision{Table~\ref{table:results}}.
\new{Understanding why combining modalities improves performance requires decomposing the effect of dataset size, modality, and feature distribution, and is an area for future work (see Section~\ref{subsec:future_modality}).}

\minihead{Effect of training method}
\revision{Existing cross-modal techniques assume one-to-one mappings between modalities~\cite{mml1,mml2,crossmodalws,mmdbms1}, whereas we have different entities for each modality and use our common feature space to construct many-to-many mappings.
To determine if general cross-modal optimizations~\cite{mmdbms1} can be leveraged in conjunction with our techniques, we evaluate if training a CNN (inception-v3~\cite{inception}) to materialize features performs better than our services. We found that our features perform up to 1.54$\times$ times better, and our proprietary (black-box, Google-wide) embedding also outperforms this generated embedding by a small (1.04$\times$) factor}.
Thus, we instead evaluate methods to fuse the two data modalities under our feature space. 
We found that the early fusion model outperformed our alternatives.
Compared to intermediate fusion, early fusion performs up to 1.22$\times$ (average 1.08$\times$) better across the tasks.
Compared to DeViSE, early fusion performs up to 5.52$\times$ (average 2.21$\times$) better.
\revision{While DeViSE represents a canonical baseline for cross-modal workloads, we hypothesize that re-using an embedding space specific to the old data modalities fails to sufficiently leverage information from the new modality.
Thus, using methods for domain adaptation~\cite{domainadaptation1} with our methods may further boost performance over these multi-modal and fusion techniques.}
This hypothesis is substantiated by recent work on understanding when and how information transfers in multi-task learning~\cite{multimodal_xfer}, which we seek to extend to our scenario. 

\eat{Thus, further evaluating how to more effectively tune our multi-modal architecture to account for modality covariance may improve intermediate fusion performance.}

\subsection{Weak Supervision in Practice}
\label{subsec:eval_ws}
In this section, we evaluate our training data curation phase. 
Correspondingly, we also look at canonical metrics for evaluating the weak supervision generative model: precision, recall and F1 score.
We first compare automatically generated labeling functions to human-generated ones, and then evaluate using label propagation as an LF.

\subsubsection{Automatic vs. Manual LF Generation}
\label{subsec:eval_ws_human}
To evaluate our automatically generated LFs, for comparison, our ground truth collection team manually developed candidate LFs for CT 1.
As previously noted, a limitation in using humans to generate LFs is that experts for each language and region are required to construct high-quality LFs.
As a result, in this subsection, we restrict our data points to English for a representative comparison.

We compare LFs in terms of time to construct and the performance of Snorkel's generative model.
The automatically generated LFs require 3.75 hours (14 minutes for itemset mining and 3.75 hours for label propagation in parallel), and the human generated LFs require 7 hours spread over days to weeks.
While both tasks can be parallelized based on machine and labeling resources, these gains are representative of typical resource availability. Despite \revision{their simplicity}, our LFs outperform the \revision{more complex, multi-feature}, human-generated LFs by 2.7 F1 points, reflecting a 14.3\% precision increase and 9.6\% recall decrease, with a 3\% coverage increase and 1.35$\times$ AUPRC improvement.
\revision{We hypothesize this reflects the discriminative power of our extracted features for CT 1, and are evaluating additional tasks to verify.}

\subsubsection{Label Propagation}
\label{subsec:eval_lp}

In \revision{Table~\ref{table:expander}}, we compare the precision, recall and F1 score of Snorkel's generative model, and the AUPRC of the discriminative model when using LFs developed with and without label propagation. 
All values show the relative improvement that label propagation provides compared to LFs generated with only itemset mining.
As stated in Section~\ref{subsec:data_curation}, label propagation provides high recall LFs, 
resulting in net F1 improvement---up to over an order of magnitude.    

In tasks such as CT 2, our automatically mined LFs are sufficient in capturing both high precision and recall, indicating the positive class is ``easier" to identify.
We can identify such cases a priori by evaluating results of WS using only the mined LFs with a text development set, saving 3.5 to 5 hours of processing time. 
Simultaneously, there are tasks such as CT 1 or CT 5 where improvements in F1 score do not translate to AUPRC improvement in the end model.
Rather than being a limitation of the method, we believe this is a limitation of the human-curated test set.
In small-scale experiments, we have verified that label propagation's improved recall signifies that it is better identifying borderline positive and negatives, and thus may be uncovering examples that were either not sampled for review (e.g., in the rare events case) or incorrectly labeled by human reviewers, but are not reflected in our final test set.
Thus, we are exploring when graph-based, nearest-neighbor methods using \orgress can de-noise, label, or identify candidate examples to label (e.g., active learning) in isolation.

\begin{table}[]
\begin{center}
\caption{Relative improvement gained in the training data curation step from using label propagation. Label propagation results in a net F1 score increase.}
\label{table:expander}
\begin{tabular}{lrrrr}
\hline
Task                  & Precision           & Recall            & F1                &    AUPRC     \\ \hline
CT 1          & 0.95$\times$        & 1.23$\times$      & 1.10$\times$      &  1.01$\times$    \\
CT 2        & 1.00$\times$       & 1.00$\times$     & 1.00$\times$     &   1.00$\times$       \\
CT 3      & 0.87$\times$        & 1.31$\times$      & 1.21$\times$      &   \ch{1.25}$\times$      \\ 
CT 4       & 1.45$\times$        & 162$\times$       & 129$\times$       &    \ch{1.24}$\times$      \\ 
CT 5       & 1.40$\times$         & 46.0$\times$        & 44.0$\times$        &    1.05$\times$      \\ \hline
\end{tabular}
\end{center}
\end{table}

\section{Discussion}

In this section, we describe avenues for future work across each of the  \splitarch steps, and comment on how to determine if a cross-modal approach should be deployed.

\subsection{Feature Generation}
\label{subsec:future_feat}
\Orgress are becoming increasingly common across organizations as teams rely more heavily on ML~\cite{OR_pinterest,OR_airbnb_ranking, OR_fair_languagemodeling}.
We demonstrate that leveraging these auxiliary resources provides opportunity to train better models for related tasks of new modalities. 

However, as the number of available resources rises, it becomes challenging to discover and curate which may be useful for a new task.
\revision{Low quality \orgress incorrectly handled may hurt model performance, but our manual curation of these sources limits our scalability to new, non-Google domains.
Methods for feature attribution would enable us to evaluate the contribution of specific data modalities and resources on a per-service basis, and developing methods to scale these techniques to our data volumes would improve performance~\cite{feature1,feature2,feature3}}.
Simultaneously, we require systems and algorithms that build off those in the data management community~\cite{modeldb,drybell,mmdbms1} to enable better service discovery and exploration.

\subsection{Training Data Curation}
\label{subsec:future_labeling}
We present an automated WS pipeline that outperforms human curated LFs. 
\revision{However, we hypothesize this is a result of our feature space easily discriminating between the two classes}.
A more robust interface would instead target expert-attention when they explore data slices and validate LF performance.
Maintaining a human in the loop may avoid potentially unwanted features emphasized in LFs that may either be unrelated to the target task or bias the training dataset in negative ways. 
A first step in achieving this is to present our mined results as a starting point for experts' exploration.
\revision{Moving forward, we hope to augment and draw from the experience of~\cite{snorkel} to understand ways to interface with non-experts to develop LFs in a cross-modal setting.}

In addition, using label propagation for data curation revealed that existing, traditional random-, rule-, and active-learning-based sampling to identify positive examples may not capture certain behavioral modes and edge-cases.
Semi-supervised methods such as label propagation provide opportunity to augment and improve existing training datasets by leveraging \orgress such as pre-trained models and corresponding embeddings.
Further understanding when semi-supervised methods augment or \emph{outperform} existing techniques in this space is an exciting area for continued research in training data management for ML~\cite{snorkel}.

\subsection{Model Training}
\label{subsec:future_modality}
We demonstrated the power of combining data modalities \revision{under an induced common feature representation, unlike previous work that operates on the  disjoint feature spaces but common label spaces}. However, this can lead to challenges. First, as the number of data modalities increases, their differences in feature distributions may compound to negatively impact performance.
Second, as the number of data modalities grows, fewer features may overlap with one another, resulting in non-uniform relationships across modalities.
As a result, certain data and label sources will be of a higher quality for a task than others.
Better managing and leveraging the relationships between data modalities, label sources, and the chosen \orgress will more effectively leverage these data sources. 
\revision{Thus, we are exploring domain adaptation as a primitive to help balance between the data modalities under our common feature space.}

\subsection{Cross-Modal vs. Fully Supervised}
\label{subsec:operating_regime}
We see in Section~\ref{subsec:eval_e2e} that there are regimes where a cross-modal approach is better than a fully supervised approach, ranging from 4k to 750k fully supervised data points.
Knowing the current operating regime in production is often unclear.
Further, offline metrics such as precision and recall on a hold out set rarely reflect production performance or business needs (i.e., relative sensitivity or user-impact of tasks).

A solution is to train and deploy models in parallel.
However, to (1) understand when models are performing poorly in production, or (2) compare the performance of many candidate models, sampling and human reviewing is often required. 
In the former, data samples---a combination of random and importance sampling---can be used to periodically check live model performance.
In the latter, the same can be performed over the classification differences between models.
Developing methods to characterize the difference between online and offline metrics across task and modalities will improve performance by understanding the operating regime.

\section{Related Work}
\label{sec:relwork}

\minihead{End-to-End Methods in ML}
Cross-modal adaptation is a form of transductive transfer learning~\cite{transferlearning1, transductive1}. 
Most work in this space assumes the same feature space across modality~\cite{transductive2,transductive3}, but with different distributions, similar to domain adaptation~\cite{domainadaptation1,domainadaptation2,domainadaptation3, domainadaptation4}.
However, our feature spaces differ.

In few- or one-shot learning~\cite{fs1,fs2}, a classifier is trained to perform well on classes with few examples.
This area focuses on learning from existing labeled data for closely related tasks/classes of a given modality~\cite{fsncm1,fsncm2,fsncm3,fsncm4}, which does not exist to begin with in our setting.
In zero-shot learning, labels of the target class do not exist at training time~\cite{zs1}.
Most closely related to our setting is work that classifies descriptions or metadata about data in the source modality~\cite{zsncm1}, and work that uses semantic representations or human-provided descriptions from other modalities to classify the target modality~\cite{zscm1,zscm2,zscm3}.
In multi-modal learning~\cite{mml1,mml2}, common feature representations represent general ``concepts" jointly across multi-modal data. 
However, a subset of tasks must still be trained in the target modality to enable mapping data points across modalities.

Weak supervision~\cite{snorkel} allows domain experts to label large volumes of data using weak, noisy labeling sources.
While weak supervision has demonstrated success in cross-modal settings for medical applications~\cite{crossmodalws}, data points are directly linked between data modalities (e.g., clinical notes and lab results), which is not true in our setting.
Upon construction of a common feature space, while we focus on the opportunities of weak supervision, we can also consider methods for unsupervised domain adaptation~\cite{unsupervisedda}, extensions to our initial integration with semi-supervised learning~\cite{ssl}, and extensions of universal or semantic embedding construction~\cite{huse,emb1,emb2,devise}.

\minihead{Multi-Modal Analytics and DBMSs}
Multi-modal content retrieval~\cite{mmdbms3,mmdbms4,mmdbms5} explores how to store, index, and query multi-modal data~\cite{mmdbms2}.
This work explores how to represent unstructured data via structured representations, as we do in our feature generation step. 
More closely related is work that focuses on transferring features across domains for multi-modal analytics~\cite{mmdbms1}, where image features derived from neural network models are used to join images with \revision{their} structured features. 
We focus on how to leverage these resources to develop models for these new modalities, and are complementary with systems such as \textsc{Vista}~\cite{mmdbms1}.

\minihead{Data Integration, Fusion, and Explanation}
We leverage methods used for attention prioritization~\cite{cidrmb} in result explanation~\cite{macrobase} and difference detection~\cite{diff} to construct a pipeline to automatically develop LFs. 
Our work is related to problems in data integration and management~\cite{dataintegration1,dataintegration2,dataintegration3}, such as data fusion~\cite{datafusion,slimfast}, in that we consider the problem of handling and managing heterogeneous data sources with widely different properties and quality to tackle a common problem of cross-modal adaptation.

\section{Conclusion}
\label{sec:conclusion}
\balance
We develop a scalable pipeline for cross-modal adaptation in production that uses \orgress to connect new and existing data modalities. 
We demonstrate how to bootstrap models for cross-modal adaptation that perform as well as a fully trained model with several thousand data points in days \revision{to weeks}, instead of months.
These results highlight the opportunity in leveraging new resources both within an organization and externally, which only become increasingly available as ML becomes commoditized.



\subsection*{Acknowledgments}
{
\small
We thank Sunita Varma, Adam Juda, Ramakrishnan Srikant, Shilpa Jain, Brianna Kaufman, Erin Rusaw, Umut Oztok, and Matthew Keegan for their support in preparing the manuscript and experimental infrastructure.  
We thank Paroma Varma, Fred Sala, Edward Gan, Kexin Rong, Daniel Kang, and members of the Stanford Infolab for their feedback. 
This research was supported in part by affiliate members and supporters of the Stanford DAWN project (Ant Financial, Facebook, Google, and VMware), Toyota Research Institute (TRI), Northrop Grumman, Amazon Web Services, Cisco, and NSF Graduate Research Fellowship grant DGE-1656518. Any opinions, findings, and conclusions expressed in this material are
those of the authors, and do not reflect the views of the NSF. TRI provided funds to assist the authors but this article solely reflects the opinions and conclusions of its authors, and not TRI or any other Toyota entity.
}

\bibliographystyle{abbrv}
\bibliography{paper}

\end{document}